\documentclass[sigconf]{acmart}

\AtBeginDocument{%
  \providecommand\BibTeX{{%
    \normalfont B\kern-0.5em{\scshape i\kern-0.25em b}\kern-0.8em\TeX}}}

\setcopyright{acmcopyright}
\copyrightyear{2018}
\acmYear{2018}
\acmDOI{XXXXXXX.XXXXXXX}

\acmConference[Conference acronym 'XX]{Make sure to enter the correct
  conference title from your rights confirmation email}{June 03--05,
  2018}{Woodstock, NY}
\acmPrice{15.00}
\acmISBN{978-1-4503-XXXX-X/18/06}

\usepackage{microtype}
\usepackage{graphicx}
\usepackage{subfigure}
\usepackage{booktabs} 
\usepackage{hyperref}
\usepackage{url}
\usepackage{multirow}
\usepackage{wrapfig}
\usepackage{float}
\usepackage{xcolor}
\usepackage{amstext}
\usepackage{amsmath}

\usepackage{amssymb}
\usepackage{mathtools}
\usepackage{amsthm}
\usepackage[capitalize,noabbrev]{cleveref}

\theoremstyle{plain}

\theoremstyle{definition}

\theoremstyle{remark}

\usepackage[textsize=tiny]{todonotes}
\newcommand{\stitle}[1]{\vspace{1ex}\noindent\textup{\textbf{#1}}}

\begin{document}

\title{Biological Factor Regulatory Neural Network}


\author{Xinnan Dai}
\authornote{The work was done during the author's  internship with Microsoft.}
\email{daixn@shanghaitech.edu.cn}
\affiliation{%
  \institution{ShanghaiTech University}
  \country{}
}

\author{Caihua Shan}
\email{caihuashan@microsoft.com}
\affiliation{%
  \institution{Microsoft Research Asia}
  \country{}
}

\author{Jie Zheng}
\email{zhengjie@shanghaitech.edu.cn}
\affiliation{%
  \institution{ShanghaiTech University}
  \country{}
}

\author{Xiaoxiao Li}
\email{xiaoxiao.li@ece.ubc.ca}
\affiliation{%
  \institution{University of British Columbia}
  \country{}
}

\author{Dongsheng Li}
\email{dongsheng.li@microsoft.com}
\affiliation{%
  \institution{Microsoft Research Asia}
  \country{}
}

\renewcommand{\shortauthors}{Dai, et al.}

\begin{abstract}
Genes are fundamental for analyzing biological systems and many recent works proposed to utilize gene expression for various biological tasks by deep learning models. Despite their promising performance, it is hard for deep neural networks to provide biological insights for humans due to their black-box nature. Recently, some works integrated biological knowledge with neural networks to improve the transparency and performance of their models. However, these methods can only incorporate partial biological knowledge, leading to suboptimal performance. In this paper, we propose the Biological Factor Regulatory Neural Network (BFReg-NN), a generic framework to model relations among biological factors in cell systems. BFReg-NN starts from gene expression data and is capable of merging most existing biological knowledge into the model, including the regulatory relations among genes or proteins (e.g., gene regulatory networks (GRN), protein-protein interaction networks (PPI)) and the hierarchical relations among genes, proteins and pathways (e.g., several genes/proteins are contained in a pathway). Moreover, BFReg-NN also has the ability to provide new biologically meaningful insights because of its white-box characteristics. Experimental results on different gene expression-based tasks verify the superiority of BFReg-NN compared with baselines. Our case studies also show that the key insights found by BFReg-NN are consistent with the biological literature.
\end{abstract}




\maketitle

\section{Introduction}

Understanding how cells work is an essential problem in biology, and it is also very important in biomedical areas because of disease phenotype and precision medicine. From a genome-scale view, the whole cell system is modeled by level, starting from DNA, mRNA, and protein to metabolomics, 
and finally, inferring the phenotype. We define these molecules and molecule sets as biological factors.
At each level, the same type of biological factors interact or regulate each other, which determines cell fate, driving the cells to develop, differentiate, and do other activities \citep{angione2019human}. 

Thanks to single-cell sequencing technologies, we can obtain gene expression data from the mRNA level, which is fundamental to analyzing the whole cell system.
Currently, gene expression data is widely used to identify cell states during cell development, characterize specific tissues or organs, and analyze patient-specific drug responses \citep{paik2020single}.

Many deep learning methods are proposed to utilize gene expression data for predictions, and achieve extraordinary performance in different biological tasks. For instance, gene expression could be treated as a type of input feature to classify cell types, cluster cells and even calculate patient survival time \citep{erfanian2021deep,huang2020deep}. 
Although most deep neural networks (DNNs) model could diagnose cancers with high precision, the original DNNs cannot tell us detailed biological factors/processes which cause cancers. 
For instance, the regulation between gene PFKL and HIF1A under HEPG2 pathway has a high probability of causing liver cancer~\citep{shoemaker2006nci60,garcia2019benchmark}.

Recently, some works leverage existing biological knowledge as graphs to represent the relations of biological factors into the prediction models, and significantly improve the prediction accuracy of specific tasks. For example, 
\cite{rhee2017hybrid} and \cite{chereda2021explaining} mapped gene expression data into the protein-protein interaction network, and used graph neural networks to predict cancer. \cite{elmarakeby2021biologically} modeled the relations of gene-pathway and pathway-biological process as a network, and used a deep neural network to diagnose prostate cancer. \cite{yu2016translation,ma2018using} used the Gene Ontology (GO) knowledgebase to build the neural network architecture, but they are too sketchy to simulate the gene or protein reactions in the cells, and may lead to suboptimal performance. Although they mitigate the black-box issues, they only use partial biological knowledge, and they cannot explore new knowledge from gene expression data. 

In this paper, we propose a generic framework, named biological factor regulatory neural network (BFReg-NN), 
whose goal is to simulate the complex biological processes in a cell system, understand the functions of genes or proteins, and ultimately give insights into the mechanism of larger living systems. Particularly, BFReg-NN is a neural network with the following architecture. 
For one thing, each neuron is mapped into a biological factor (e.g., a specific gene or protein), and arranged level by level based on the hierarchy of biological concepts, such as genes, proteins, pathways, biological processes, and so on. For another, since biological factors regulate each other, edges between neurons (and hyperedges among neurons) are set to reflect the existence of these regulations. In such a manner, edges also model biological meanings.  

Further, we apply two different operations to simulate all the types of interactions in BFReg-NN. Inside one layer, genes regulate each other and create feedback loops to form cyclic chains of dependencies in the regulatory network. Therefore, graph neural network styled operations are suitable to build the gene reactions in the k-hop neighborhood, and obtain the ``steady state" of genes. It is the same for proteins in PPI. In the layer of pathways, it can be regarded as a hypergraph where each hyperedge is a pathway containing multiple proteins. Accordingly, the hypergraph neural network is used to aggregate and balance the information of each pathway. Across the layers, we need to imitate the material transformation (e.g., genes translate protein). Thus, we adopt deep neural network styled operations to map the relations.

\begin{figure*}[t!]
    \centering
\includegraphics[scale=0.5]{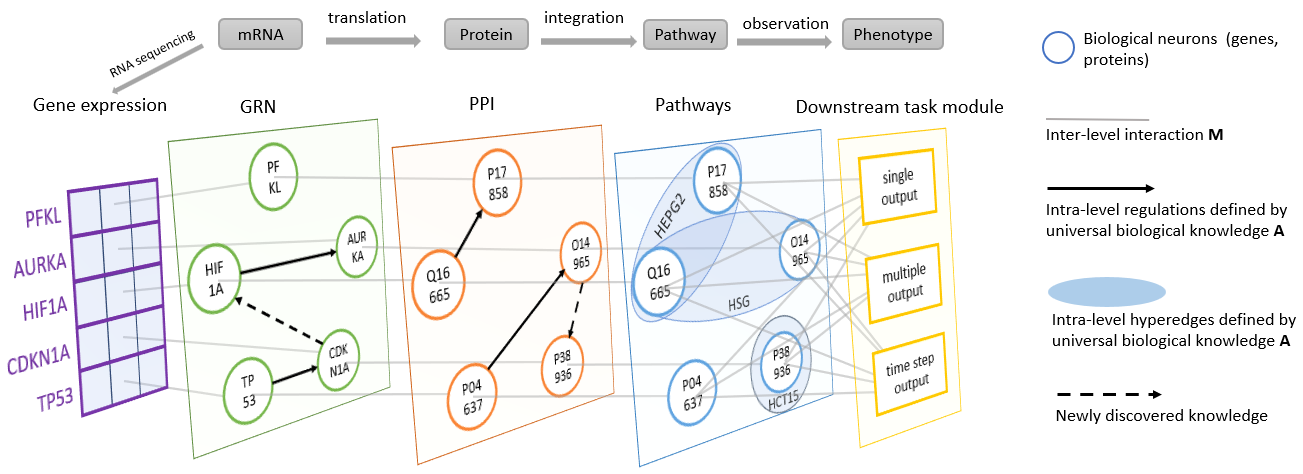}
    \caption{The pipeline of BFReg-NN. We build the hierarchical biological network inspired by the cell system, modeling several levels to separate mRNA, protein, pathway and phenotype, and simulating biological factor interactions both at the intra-level and inter-level. We obtain the gene expression value from RNA sequencing as input, to predict the property of the phenotype as output. Taking the cancer dataset NCI-60 as an example. We know that the regulatory activations start from gene HIF1A and TP53 to gene AURKA and CDKN1A at the gene level \citep{garcia2019benchmark}. And at the protein level, the translated proteins Q16665 and P04637 stimulate P17858 and O14965 respectively \citep{turei2016omnipath}. We mark those relations as existing biological knowledge by the black solid lines. Multiple genes with their products consist of pathways to drive cells to different types. For example, PFKL and HIF1A are activated in the HEPG2 cell line pathway, which leads to a type of liver cancer \citep{shoemaker2006nci60}.  Besides, new regulatory relations would be identified at each level shown as dotted lines. After training, we learn the good representations of biological factors, and employ them for different downstream tasks with various output formats. 
}
    \label{method_fig}
\end{figure*}

Moreover, BFReg-NN is flexible to explore new biological knowledge by trying to add non-existing edges in the network. The reason is that some edges are still hard to be discovered due to technological limitations and rewiring phenomenons in individual cells. If the prediction performance is improved by added edges, it is more likely that these edges exist. We illustrate BFReg-NN simulates on genome-scale cell system as an example in Figure~\ref{method_fig}.

The advantages of our proposed model include: (1) Compared with previous works, BFReg-NN merges with the structural biological knowledge in cell systems, including hierarchical relations (e.g., genes-proteins, proteins-pathways mappings), and regulatory relations among certain factors, such as GRN and PPI. Therefore, it could imitate how different biological factors work inside a cell. 
(2) The model of BFReg-NN is transparent and interpretable, as each neuron and edge has its corresponding biological meaning. Thus, the learned model weights give evidence of which biological parts are activated and which biological products are generated, leading to the final prediction. (3) By adding new edges between neurons, BFReg-NN not only achieves better performance in downstream tasks, but also has the potential to complete undiscovered biological knowledge. Traditional knowledge completion methods for biological domains (e.g., link prediction by knowledge bases/graphs) suffer from imbalanced data problems \citep{bonner2021implications}.  
BFReg-NN utilizes the gene expression data, which reflects the real cell states, and thus obtains more reliable results.

In the experiment, we show the effectiveness of BFReg-NN on several kinds of biological tasks, including missing gene expression value prediction, cell classification, future gene expression value forecasting and cell trajectory simulation. We also test the knowledge completing ability of BFReg-NN by the recall of the existing biological knowledge. Finally, we do case studies for newly discovered knowledge. The results demonstrate that BFReg-NN provides biologically meaningful insights. 

\section{Related Work}

\stitle{Gene expression and its applications:}
By RNA sequencing, it is easy to obtain gene expression which is a value to represent the amount of gene transcripts from a DNA fragment~\cite{eberwine2014promise}. It has been used in a variety of biological applications, including single-cell analysis~\citep{yu2022zinb,zhou2022kratos}, disease diagnosis~\citep{xing2022multi} and drug discovery~\citep{pham2021deep}. 
But most of these models lack transparency and ignore the existing biological knowledge. 

\stitle{Knowledge graph enhanced downstream tasks:}
The emergence of knowledge bases/graphs has led to enhancing the performance in many fields of computer science, such as computer vision and natural language processing~\citep{ren2021lego, hao2021ks, liu2021auto}.  
Similarly, knowledge graphs also have been widely used for specific biological tasks such as cancer diagnosis in recent years~\citep{elmarakeby2021biologically,rhee2017hybrid}.
Some methods \cite{ma2018using,yu2016translation,zhang2022ontoprotein} used the Gene Ontology (GO) knowledge, which defines GO terms (e.g., molecular function, cellular component, biological process) and builds the neural network architecture based on term relations. However, the network is too sketchy to simulate the complex gene/protein reactions in the cell, leading to suboptimal performance.
OntoProtein~\citep{zhang2022ontoprotein} embedded  the gene ontology knowledge in pre-training to improve the performance of several protein downstream tasks. But it limited the interpretation ability because of transforming knowledge into embeddings. 
P-Net~\cite{elmarakeby2021biologically} modeled the relations of gene-pathway and pathway-biological process as a neural network to diagnose prostate cancer. 
GLRP~\cite{chereda2021explaining} assigned genes to the PPI network to predict breast cancer, and then explained the important genes by Layer-wise Relevance Propagation. 
Although they mitigate the black-box issues, they all use a small part of biological knowledge, and they cannot explore new knowledge from gene expression data.

\stitle{Knowledge complement:}
Our work is related to knowledge complement~\citep{yao2019kg,goel2020diachronic}. In particular, lots of work on biological knowledge complement has been done recently~\citep{yu2021sumgnn,zitnik2018modeling}. Much attention is paid to predicting the relation between specific biological factors, which can be regarded as a link prediction problem. For example, 
\cite{fout2017protein} classified whether ligand and receptor proteins interact based on their molecular structures. 
\cite{mohamed2020discovering} predicted drug-target interaction (DTI) by learning representations of drugs and targets from the KEGG database. 
\cite{hamilton2018embedding} discovered new drugs for diseases by embedding the drug-gene-disease database in multiple relations. However, existing biological knowledge is collected in imbalance, which causes little biological meaningful predictions~\citep{bonner2021implications}. In our work, we expect discovered knowledge from gene expression data, which can interpret the cell state by biological factor interactions and avoid imbalanced situations.

\vspace{0.7em}
\section{Biological Background}

The biological system is modeled by a complex network that connects many biologically relevant entities to work together to perform one or more particular functions. It could be at the organ/tissue scale, such as the nervous system, or the integumentary system. On the micro/nanoscopic scale, examples include cells, organelles, and so on. In this work, we focus on the simulation of the biological system in a cell at a genome-scale.

\newpage

Thanks to the development of sequencing technologies, it is easy and cheap to obtain an amount of gene expression data to build genome-scale analysis. Starting from genes, we model the cell genome-scale system by single-omic level (intra-level) regulations and different omic level (inter-level) mappings \citep{angione2019human}. Among different omic levels, defined as inter-level in BFReg-NN, genes are first transcribed from DNA and then translate to proteins. Then, pathways integrate individual genes or protein products to perform certain cell functions. To model single-omic level regulations, we define the intra-level interactions. Genes are regulated by each other, known as GRN, which means the gene expression values are governed to be inhabited or activated by association molecule products, such as RNA. The interactions among proteins are similar, where they inhabit, activate, or combine with others to influence the expression values (or protein abundances) in cells, called PPI. Notice that in general, the relations among factors are universal for all the cells, thus the biological knowledge could adapt to different types of cells. But sometimes the 
biological factor interaction would rewire in specific cells, and thus the extra knowledge discovered and verified is also important \citep{lynch2011transposon}.

\section{Methodology}

Assume we have gene expression data $\mathbf{x}\in \mathbb{R}^{n}$ show the state of a cell, where $n$ is the number of genes. Besides, we also define the neural network architecture of BFReg-NN according to structural biological knowledge with the intra-level regulations $\mathbf{\mathcal{A}}$ and the inter-level mappings $\mathbf{\mathcal{M}}$. After training BFReg-NN, we obtain the hidden embedding $\mathbf{H}_{i}= \text{BFReg-NN}(\mathbf{x}, \mathbf{\mathcal{A}}, \mathbf{\mathcal{M}})$ for each gene $i$, and do predictions for downstream tasks. 

The output format of predictions could be a single value $y=f(\mathbf{H})$, or multiple values $\mathbf{\bar{x}}=f(\mathbf{H})$.
Further, we also handle the time-series output. Given the gene expressions of cells at the $t_0$ time, we aim to predict the gene expressions $\mathbf{\hat{X}}\in\mathbb{R}^{n\times T}=f(\mathbf{H})$ in the following $t_1,\dots,T$ time steps. Finally, BFReg-NN could explore new insights by adding more edges $\mathbf{\mathcal{A}'}$. 

In the following sections, we first extract $\mathbf{\mathcal{A}}$ and $\mathbf{\mathcal{M}}$ from the existing biological databases/graphs. Then, we propose two versions of BFReg-NN. Finally, we introduce how to apply BFReg-NN to different downstream tasks. 

\subsection{Biological knowledge databases/graphs}
Based on the existing biological knowledge, we first divide biological factors into different levels, $\mathcal{L}=\{\text{Gene}, \text{Protein}, \text{Pathway}, ...\}$. At each level, regulatory relations between factors in the intra-level are formulated as a matrix set $\mathbf{\mathcal{A}}=\{\mathbf{A}_{\text{Gene}},\mathbf{A}_{\text{Protein}}, \mathbf{A}_{\text{Pathway}}, $ $\dots,\mathbf{A}_{L}\}$, where $\mathbf{A}_{\text{Gene}}$ could be gene relations and defined by GRN, and $\mathbf{A}_{\text{Protein}}$ is decided by PPI. 
$\mathbf{A}_{\text{Pathway}}$ is a little special because it is a hypergraph, where each edge could connect more than two nodes, called the hyperedge. The proteins in a hyperedge propagate the information to influence each other directly.

The values in $\mathbf{A}_{l}$ are binary to represent the existence of relations. We also identify the binary mapping matrixes $\mathbf{\mathcal{M}}=\{\mathbf{M}_1,\mathbf{M}_2,\dots,\mathbf{M}_{L-1}\}$ from level $l$ to its upper level $l+1$ as the inter-level interaction. The dimension of $\mathbf{M}_l$ is dependent on the factor numbers of the two levels. We set $\mathbf{M}_1$ to map between genes and proteins, and $\mathbf{M}_2$ as a direct mapping between protein-level and pathway-level. $\mathbf{\mathcal{A}}$ and $\mathbf{\mathcal{M}}$ both decide the architecture of the neural network, including neurons and links between neurons.  

\subsection{Basic BFReg-NN model}

Given the gene expression of a cell $\mathbf{x}$, we first utilize an embedding layer to encode each gene independently, where $\mathbf{H}^{0,0}_{i} = \text{Emb}(\mathbf{x}_i)$. The embedding layer lets genes share the same parameter MLP layer to obtain the gene expression representation. As the gene expression is in a floating form with a little measurement error, the embedding layer is utilized to reduce this error and thus enhance the performance of the model.
Then we obtain $\mathbf{H}^{0,0} \in \mathbb{R}^{n\times d}$ as the input to neurons at the first level, i.e., gene neurons. 

We have $\mathbf{A}_l$ as intra-level relations between neurons at level $l$. Inspired by graph neural networks, we use the message passing mechanism to update $\mathbf{H}^{l,k}$, which is the hidden representation of the neuron in the $k$-hop at level $l$. The formulation is:
\begin{equation} 
\mathbf{H}_{i}^{l,k+1} = \text{update}\left(\sum_{j \in \mathbf{A}_l(i)}\text{message}\left(\mathbf{H}_{i}^{l,k},\mathbf{H}_{j}^{l,k}\right), \mathbf{H}_{i}^{l,k}\right).
\end{equation}
The message function is to generate the message from neuron $j$ to neuron $i$, where $\mathbf{A}_l(i)$ decides which neurons are neighbors for neuron $i$. Then the update function is to update the embedding of neuron $i$ using the obtained messages and the previously hidden embedding. $k\in [0,K-1]$ is the number of hops to determine that neuron $i$ is influenced by other neurons in the $K$-hop neighborhood. Batch normalization is conducted at the end of each level. In detail, we utilize GAT-styled (graph attention) \citep{vaswani2017attention} to compute the attention weight for each message, and weighted sum up all the messages to update. 
As for operations in hypergraphs, we apply HGNN-styled (hypergraph neural network)~\citep{bai2021hypergraph} message and update functions. 
Specifically, the hypergraph can be represented by an incidence matrix $\mathbf{R} \in \mathbb{R}^{|V_{\mathbf{R}}|{\times}|E_{\mathbf{R}}|}$, where $\mathbf{R}_{ij}=1$ means the node $i$ belongs to the hyperedge $j$ (i.e., the protein $i$ is contained in the pathway $j$). The equation is 
$$
\mathbf{H}^{l,k+1} = \sigma( \mathbf{D}_\mathbf{R}^{-1} \mathbf{R} \mathbf{B}_\mathbf{R}^{-1} \mathbf{R}^\mathsf{T} \mathbf{H}^{l,k} \mathbf{W}^{l,k} )
$$
where $\mathbf{D}_\mathbf{R}$ and $\mathbf{B}_\mathbf{R}$ are two corresponding degree matrices for normalization, and $\mathbf{W}^{l,k}$ is the learnable matrix.  

Due to multiple-level interactions in biological systems, the embedding is also learned level by level. Since $\mathbf{M}_l$ is inter-level relations between level $l$ and $l+1$, we utilize the masked deep neural network to update the initial representation $\mathbf{H}^{l+1,0}$ at next level 
\begin{equation} 
\mathbf{H}^{l+1,0} = \text{activation}\left( \left(\mathbf{M}_l \odot \mathbf{W}^{l,K} \right) \mathbf{H}^{l,K} + \mathbf{b}^{l}\right).
\end{equation}
$\mathbf{H}^{l,K}$ is the output after batch normalization. The element-wise multiplication $\mathbf{M}_l \odot \mathbf{W}^{l,K}$ ensures that non-existing relations are not used for updating. 
$\mathbf{W}^{l,K}$ and $\mathbf{b}^{l}$ are learnable parameters. 

\subsection{Enhanced BFReg-NN model}

Here the enhanced version is introduced by adding new edges in $\mathcal{A}=\{\mathbf{A}_{\text{Gene}}, \mathbf{A}_{\text{Protein}},$$\mathbf{A}_{\text{Pathway}},$$ ...\}$. It can explore new biological insights and improve performance simultaneously. 

Existing biological knowledge is detected by biological technology to reflect the implicit relations among factors. However, some knowledge is still hard to be discovered due to technological limitations and rewiring phenomenons in individual cells. Therefore, we spilt the interaction into two types. One is the universal regulation, supported by existing knowledge $\mathcal{A}$. The other is local interaction, inferring the new biological knowledge or rewiring in individual cells, but now hidden in the non-existent edges of $\mathcal{A}$.  
Instead of a binary matrix $\mathbf{A}_l$ used in the basic model, we use $\mathbf{A}_l$ to constrain the learnable matrix $\mathbf{A'}_l$ to discover new knowledge.
As universal regulations are verified by biological methods, we use them the same as the basic model. For non-existent edges, we reweigh it by a small value $0\leq\alpha<1$ due to it being less convincing. Thus, the edge intensity based on two types of knowledge is modified as:
\begin{equation}
\mathbf{A'}_{l}=\begin{cases}
\omega_{ij}^{l}, \quad \text{ universal regulation that already exists in $\mathbf{A}_{l}$ } \\
\alpha \omega_{ij}^{l}, \quad \text{ local interaction that is ignored in  $\mathbf{A}_{l}$ }
\end{cases}
\end{equation}
where $\omega_{ij}^{l}=\sigma(\text{MLP}(\text{concat}[\mathbf{H}_i^{l},\mathbf{H}_j^{l}])).$
In the enhanced model, we not only learn the neuron embeddings, but also utilize these embeddings with an MLP transformation to infer the intensity of the hidden interaction between neuron $i$ and $j$. Then we update the representations by 
\begin{equation} 
\mathbf{H}^{l,k+1} = \sigma \left(\mathbf{A'}_l \mathbf{H}^{l,k} \mathbf{W}^{l,k}  \right).
\end{equation}

After the model is trained to converge, we obtain the learned weights for non-existence edges, which provides insights for new knowledge and the rewiring phenomenon. We sort the weights $w_{ij}^{l}$ to identify the candidates which deserve verification by biological experiments. Here we use a simple edge weight modification method instead of gated edges implemented by the gumble-softmax function or advanced graph structure learning algorithms~\citep{zheng2020robust,jin2020graph}. The reason is that the biological knowledge is not sparse, and even dense in some core genes. Thus, it does not satisfy the sparse and low-rank requirements. Besides, we also implement a neural network where the existence of new edges is gated by the gumble-softmax function, but it does not work in the experiments. 
Our simple method does not add much extra computation cost while achieving great effectiveness in biological tasks. 

\subsection{Downstream Tasks}
After we obtain the final embedding $\mathbf{H}^{L,K}$, we could conduct different types of downstream tasks, whose output format could be one-dimensional, multiple-dimensional, or time-series. Here we illustrate them with three specific tasks, missing gene expression value prediction, future gene expression value forecasting, and cell classification.

\stitle{Missing gene expression value prediction:} Suppose we have a gene expression vector  $\mathbf{x}\in\mathbb{R}^{n}$ for $n$ genes, measured by the single-cell sequencing technology. Some parts in the vector are zeros or in a low value because of dropout events~\citep{gong2018drimpute}, causing biases among cells. So we aim to accurately recover all the missing values and obtain a new vector $\mathbf{\hat{x}}$. 
Since the gene expression is dependent on a certain cell, we merge gene representations. The new gene value vector is generated directly by $\mathbf{\hat{x}} =\text{MLP}(\mathbf{H}^{L,K})= \text{MLP}(\text{BFReg-NN}(\mathbf{x}))$. We minimize the mean squared loss between predicted values and ground truth for all the non-zero elements in $\mathbf{x}$. Finally, the missing values are imputed with the predicted values.

\stitle{Cell classification:} 
This task is to classify the cell types using gene expression values $\mathbf{x}$, which is important to determine the situation of tissues or patients.
Because cell type is the observable result of a biological system, we simulate each gene to pass the multiple levels of transformation to infer the property of the cell. The prediction is computed by $\hat{y}=\text{MLP}(\mathbf{H}^{L,K})=\text{MLP}(\text{BFReg-NN}(\mathbf{x}))$. It is a multi-class classification task and we employ a cross-entropy loss.

\stitle{Future gene expression value forecasting:} The single-cell data is efficient to analyze the cell response under different drugs.
However, it is expensive to collect the data at different time steps and draw the development trend of cells. Thus, we model the cell response by future gene expression value prediction. In other words, given the gene expression data $\mathbf{x}\in\mathbb{R}^{n}$ at $t_0$, we aim to forecast the gene expression data $\mathbf{\hat{X}}\in\mathbb{R}^{n\times T}$ in the following $t_1,\dots,T$ time steps. 
We use two backbones, MLP and LSTM. MLP predicts the gene expression in the future time steps simultaneously, and the equation is  $\mathbf{\hat{X}}=\text{MLP}(\text{BFReg-NN}(\mathbf{x}))$. LSTM models the dynamic values step by step, where it takes the last time output as the next step input, and the equation is $\mathbf{\hat{x}}^t=\text{LSTM}(\text{BFReg-NN}(\mathbf{\hat{x}}^{t-1}))$. We employ MSE as the loss function.

\section{Further Applications}
BFReg-NN has the potential to be plugged into more sophisticated models, such as generative models and pre-learning. 

\subsection{Cell trajectory simulation}
In this task, we regard cells as a population to understand the population dynamics. In other words, our goal is to simulate the evolution of the distribution of gene expression over time.
Since the single-cell sequencing technology makes the cells die after measurement, we can only know the distribution of gene expression of a part of cells at a certain time step, i.e., $\{x_{1, t_i}, x_{2, t_i}, \cdots, x_{N_i, t_i}\}$ where $i=\{1, 2, \cdots, T\}$. 
Based on previous papers~\cite{tong2020trajectorynet, yeo2021generative}, we also utilize continuous normalizing flows~\cite{chen2018neural} as the generative model to learn an underlying differentiation landscape from scRNA-seq data. The main difference is that our proposed BFreg-NN is utilized as $f_\theta$ instead of a traditional DNN. Because the architecture of BFReg-NN is built on biological factors and their interactions are also regarded as differential equations, BFReg-NN has a better inductive bias to model complex relationships and further improves the performance. 

Specifically, assume that $P_t(\boldsymbol{x})$ is the distribution of gene expression of cells at time step $t$, and $\boldsymbol{x}(t)$ is the gene expression of a cell drawn from  $P_t(\boldsymbol{x})$. Let $\frac{\mathrm{d} \boldsymbol{x}(t)}{\mathrm{d} t} = f_\theta(\boldsymbol{x}(t), t)$ be a differential equation describing a continuous-in-time gene transformation of $\boldsymbol{x}$. We seek to train $f_\theta$ satisfying
$$ \small
\begin{aligned}
&\boldsymbol{x}(t_i) \sim P_{t_i}(\boldsymbol{x}) \quad i=0, 1, 2,\cdots, T \\
&\frac{\mathrm{d} \boldsymbol{x}(t)}{\mathrm{d} t} = f_\theta(\boldsymbol{x}(t), t) \text{   for } t \in [t_0, t_T].
\end{aligned}
$$
According to the characteristic of continuous normalizing flows, we have 
$$
\log P_{t_T}(\boldsymbol{x}(t_T)) = \log P_{t_0}(\boldsymbol{x}(t_0)) - 
\int_{t_0}^{t_T} \text{Tr}( \frac{\partial f_\theta(\boldsymbol{x}(t), t ) }{\partial \boldsymbol{x}(t)}) \mathrm{d}t.
$$ 
Therefore, given a batch of empirical samples $\{x_{0, t_i}, x_{1, t_i}, \cdots, x_{N_i, t_i}\}$ where $i=\{0, 1, \cdots, T\}$, we can first map the distribution at the time $t_i$
into the distribution at the time $t_{i-1}$. Then we compare the distribution difference to make it as the same as possible. We use the wasserstein distance as the loss to evaluate the difference. 


Then we utilize  BFReg-NN to instantiate $f_\theta$. Assume that we only have one layer, such as gene layer. We first embed $\boldsymbol{x}$ by $\mathbf{H}^{0} = \text{Emb}(x)$, and change the message passing for genes 
\begin{equation} \small
\mathbf{H}^{k+1} = \mathbf{H}^{k}+\int_{t_i}^{t_i+\Delta_t} f_\theta(\mathbf{H}(t),t)	\mathrm{d}t 
\end{equation}
where $f_\theta(\mathbf{H}(t),t) = \sigma(A' \mathbf{H}(t) \mathbf{W}(t)) \mathbf{U}(t)$, and $\mathbf{W}(t)$ and  $\mathbf{U}(t)$ are produced by the hypernetwork~\citep{ha2016hypernetworks} to make them change over time.
When we have different $L$ layers in BRReg-NN, which mean $f_\theta$ is a piecewise function based on the value of $t$. One straightforward and effective approach is to define $f_\theta$ as:
$$ \small
 \begin{cases}
f_{\theta_1} \leftarrow \sigma(A'_1 \mathbf{H}(t) \mathbf{W}^1(t)), \text{  }  t \in [t_i, t_i+ \frac{1}{L}\Delta_t) \\
f_{\theta_2} \leftarrow \sigma((\mathbf{M}_1\odot \mathbf{W}^2(t))\mathbf{H}(t)+\mathbf{b}^2(t)), \text{  } t \in [t_i+\frac{1}{L}\Delta_t, t_i+ \frac{2}{L}\Delta_t)\\
\cdots \\
f_{\theta_L} \leftarrow \sigma(A'_L \mathbf{H}(t) \mathbf{W}^L(t)) \mathbf{U}(t), \text{  } t \in [t_i+\frac{L-1}{L}\Delta_t, t_i+ \Delta_t]. \\
\end{cases}
$$
$f_{\theta_1}$ imitates the interactions in gene layer, $f_{\theta_2}$ simulates the mapping from genes to proteins, and then the following $f_{\theta_{i}}$ imitate the reactions in higher layers.

\subsection{Pre-training}
BFReg-NN is also suitable in pre-training and fine-tuning framework. For example, we utilize the missing gene expression prediction task to pre-train a BFReg-NN. Then we freeze the parameters of BFReg-NN except for the last MLP layers, and fine-tune the last MLP layers in downstream tasks, such as forecasting future gene expression and cell classification.

\section{Experiments}
In this section, we first describe the experimental setting for each task, including datasets, evulation metrics, network architecture, baselines and training details (Section ~\ref{sec:setup}). Then we present the main results in Section~\ref{sec:static_dynamic}. The cell trajectory simulation and pre-training are conducted in Section~\ref{sec:cell_result} and Section~\ref{sec:pre-training_result}.
Further, Section~\ref{sec:knowledge_result} shows the ability of BFReg-NN to discover the new biological knowledge. 
Finally, we did some ablation study to evaluate the each part of BFReg-NN in Section~\ref{sec:ablation_result}. 
All the experiments are run on a single A100 GPU. The code and toy datasets are in https://github.com/DDigimon/BFReg.

\begin{table}[t]
\centering
\caption{Statistics for cell classification datasets}
\resizebox{0.48\textwidth}{!}{
\label{cdata}
\begin{tabular}{lcccccc}
\toprule
Dataset   & \#Sample  & \#Gene & GRN knowledge & PPI knowledge & Pathways & \#Label \\
\midrule
GSE       & 4658          & 418         & 172           & 136           & 89      & 8         \\
lung      & 1642          & 382         & 148           & 332           & 121     & 11        \\
muscle    & 1051          & 401         & 317           & 401           & 119     & 6         \\
trachea   & 3587          & 326         & 109           & 207           & 110     & 5         \\
diaphragm & 853           & 355         & 239           & 299           & 115     & 5        \\
\bottomrule
\end{tabular}
}
\vspace{-1em}
\end{table}

\subsection{Setup}
\label{sec:setup}
%

\subsubsection{Datasets} 
Here we describe different datasets for all the tasks mentioned in Section 4 and 5.

\stitle{Missing gene expression value prediction:} 
In this task, we collect 10 cell lines (BT20, HS578T, LNCAP, A549, MCF7, MCF10A, MDAMB231, PC3, SKBR3 and A375) from the L1000 dataset. Because of the limitation of sequencing technology, the L1000 dataset does not have the golden standard values for non-landmark genes. Thus, we employ the pre-processed data based on \cite{qiu2020bayesian}.
Besides, we also ignore the time point information in cell lines. In each cell line, we have 1482 cell samples and 714 genes. We also collect 541 and 2305 existing knowledge edges for GRN and PPI respectively. We randomly mask the gene expression value with a 60\% probability, and predict these missing values. The cell samples are randomly split by 60/20/20\% as training/validation/test data. The evaluation metric is mean square error (MSE), which means the smaller the better.

\stitle{Cell classification:} We gather 5 datasets to predict cell types, and data statistics are summarized in Table~\ref{cdata}. The first one is GSE756888 (breast cancer) collected from \cite{chung2017single}. The following four datasets are about organs, including muscle, diaphragm, lung and trachea, obtained from \cite{tabula2018single}.
We pre-process the data by deleting the cells and genes which have a larger ratio of zeros. The remaining data are summarized in Table \ref{cdata}. We also split the data same as the missing value task, and evaluate the results by macro AUC.

\stitle{Future gene expression value forecasting:}  For future value prediction, we choose the breast cancer dataset from 2019 Dream Challenge \citep{gabor2021cell}. There are 44 cell lines and 5 treatments. Different treatments mean using various drugs on cells to test the drug effects. So we have 220 situations. In each situation, the detailed process to obtain time-series gene expression data is: all the cells are first treated and then divided into several groups. At (ir)regular time step, a group of cells is killed and  their gene expression is measured. There are seven timestamps, and about hundreds of cells are measured at each timestamp. To avoid noise caused by this measurement on each individual cell, we calculate the median for each gene. Thus, the format of data is $\mathbf{X} \in \mathbb{R}^{{N_\text{treatment} \times N_\text{cell\_line}  \times N_\text{timestamp} \times N_\text{gene} }}$.

Since the data are quantified at the proteomic level, we have 37 biomarkers as genes and 38 related PPIs as the existing knowledge, which is a small protein-protein interactions graph where nodes are phosphorylated genes.   Different from the previous two tasks, we use 5 treatments as training data, and the remaining one as test data. The 5-fold cross-validation is done. We use MSE and Pearson correlation coefficient (PCC) to evaluate average performance.

\stitle{Cell trajectory simulation:} In this task, we use the same dataset as future gene expression forecasting. Instead of calculating the median value, we consider the evolution of the distribution of gene expression. Thus, we sample 512 cells at each timestamp. The current format of data is $\mathbf{X} \in \mathbb{R}^{{N_\text{treatment} \times N_\text{cell\_line}  \times N_\text{timestamp} \times N_\text{gene} \times N_\text{cell}}}$. In each situation (i.e., a certain combination of cell lines and treatments), we use the first two timestamps as known data, and predict the following trend for the remaining timestamps. The wessentrian distance is utilized to evaluate the similarity between two distributions. The performance is also averaged by the number of predicted timestamps.

\stitle{Pre-training:} The dataset and evaluation metric are also the same as future value forecasting.

\begin{table*}[t!]
\setlength{\tabcolsep}{2.8pt}
\centering
\caption{Results on missing gene expression value prediction (evaluated by MSE$\downarrow$).} 
\resizebox{0.95\textwidth}{!}{
\label{r_result}
\begin{tabular}{clllllllllll}
\toprule
\textbf{Knowledge}                                  & \textbf{Models}             & \textbf{BT20}           & \textbf{HS578T}         & \textbf{LNCAP}          & \textbf{A549}           & \textbf{MCF7}           & \textbf{MCF10A}         & \textbf{MDAMB231}       & \textbf{PC3}            & \textbf{SKBR3}          & \textbf{A375}           \\
\midrule
\multirow{3}{*}{\shortstack[c]{no\\knowledge}}        & MLP                & \textbf{0.314} & 0.344          & 0.344          & 0.320          & 0.326          & 0.337          & 0.325          & 0.378          & 0.317          & 0.336          \\
                                           & Transformer        & 0.396          & 0.377          & 0.406          & 0.383          & 0.396          & 0.399          & 0.418          & 0.402          & 0.387          & 0.407          \\
                                           & Gated NN           & 0.709          & 0.681          & 0.644          & 0.533          & 0.647          & 0.700          & 0.667          & 0.722          & 0.594          & 0.729          \\
\hline                                           
\multirow{2}{*}{\shortstack[c]{co-expression}} & GCN                & 0.715          & 0.714          & 0.722          & 0.653          & 0.688          & 0.675          & 0.707          & 0.731          & 0.677          & 0.702          \\
                                           & GAT                & 0.719          & 0.715          & 0.723          & 0.655          & 0.693          & 0.669          & 0.720          & 0.721          & 0.676          & 0.716          \\
\hline                                           
\multirow{6}{*}{\shortstack[c]{existing\\knowledge}}           & MLP+N2V            & 0.351          & 0.378          & 0.380          & 0.349          & 0.368          & 0.350          & 0.358          & 0.392          & 0.343          & 0.354          \\
                                           & GCN                & 0.331          & 0.329          & 0.330          & 0.304          & 0.340          & 0.313          & 0.319          & 0.338          & 0.331          & 0.317          \\
                                           & GAT                & 0.324          & 0.331          & 0.332          & 0.308          & 0.321          & 0.314          & \textbf{0.317 }         & 0.347          & 0.317          & 0.315          \\
                                           & P-NET              & 0.349          & 0.335          & 0.340          & 0.340          & 0.321          & 0.352          & 0.336          & 0.354          & 0.322          & 0.331          \\
                                           & BFReg-NN(Basic)    & 0.318          & 0.336          & 0.332          & 0.329          & \textbf{0.308} & 0.328          & 0.328          & 0.344          & 0.318          & 0.328          \\
                                           & BFReg-NN(Enhanced) & 0.318          & \textbf{0.317} & \textbf{0.320} & \textbf{0.297} & \textbf{0.308} & \textbf{0.306} & 0.320 & \textbf{0.335} & \textbf{0.307} & \textbf{0.308}\\
\bottomrule                                           
\end{tabular}
}
\end{table*}

\begin{table*}[t]
\centering
\caption{Results on cell classification (evaluated by macro AUC$\uparrow$).} 
\resizebox{0.7\textwidth}{!}{
\label{c_results}
\begin{tabular}{cllllll}
\toprule
\textbf{Knowledge}     & \textbf{Models}            & \textbf{GSE}          & \textbf{muscle}       & \textbf{diaphragm}      & \textbf{lung}         & \textbf{trachea}         \\
\midrule
\multirow{3}{*}{\shortstack[c]{no prior\\knowledge}}        & MLP                & 0.9122          & 0.8586          & 0.7881          & 0.8545          & 0.9387          \\
                                            &Random Forest      & 0.9688          & 0.8771          & 0.7161             & 0.8834          & 0.9273           \\
                                            &XGBoost            & 0.8867          & 0.8534          & 0.7549             & 0.8625          & 0.9193           \\
                                           
                                           & transformer        & 0.9476          & 0.8785          & \textbf{0.8717} & 0.8900          & 0.9242          \\
                                           & Gated NN           & 0.7910          & 0.7784          & 0.7110          & 0.8240          & 0.9120          \\
\hline                                           
\multirow{2}{*}{\shortstack[c]{co-expression}} & GCN                & 0.6916          & 0.6424          & 0.5945          & 0.5822          & 0.6612          \\
                                           & GAT                & 0.7790          & 0.6882          & 0.6305          & 0.6312          & 0.6294          \\
\hline                                           
\multirow{6}{*}{\shortstack[c]{existing\\knowledge}}           & MLP+N2V           & 0.9064          & 0.8772          & 0.7501          & 0.8180          & 0.9404          \\
                                           & GCN                & 0.9285          & 0.8449          & 0.7896          & 0.8581          & 0.9421          \\
                                           & GAT                & 0.9255          & 0.8470          & 0.8039          & 0.8246          & 0.9409          \\
                                           & P-NET              & 0.9052          & 0.8654          & 0.7973          & 0.8425          & 0.9332          \\
                                           &DCell              & 0.9482          & 0.7158          & 0.6731           & 0.7545          & 0.9322           \\
                                           & BFReg-NN(Basic)    & 0.9476          & 0.8808          & 0.8420          & 0.8808          & 0.9376          \\
                                           & BFReg-NN(Enhanced) & \textbf{0.9693} & \textbf{0.8884} & 0.8509          & \textbf{0.8903} & \textbf{0.9446}  \\
\bottomrule
\end{tabular}
}
\end{table*}

\subsubsection{Neural network architecture} 
The architecture is totally defined by existing knowledge. From open-source databases, we select Dorothea \citep{garcia2019benchmark} as GRN construction, and Omnipath \citep{turei2016omnipath} as PPI. As the pathway is specific in different cells, we collect pathways for each dataset by Enrich \citep{chen2013enrichr}, which provides knowledge of the hyperedge connection among genes and proteins, including 
LINCS\_L1000\_Ligand\_Perturbations for missing value task and WikiPathways\_2019\_MOUSE(HUMAN) for cell classification task. For future value forecasting and Cell trajectory simulation, we utilize its corresponding relations (38 related PPIs) in the dataset.

\subsubsection{Baselines}
We divide baselines into three types: \\
\noindent (1) There is no prior knowledge integrated into the model, and the input is only gene expression data.
\begin{itemize}
\item \textbf{MLP} is the simplest model for static prediction. 
\item \textbf{LSTM} is the typical RNN-based model for dynamic value prediction. 
\item \textbf{Random Forest} and \textbf{XGBoost} are two classical tree-based models. 

\item \textbf{Transformer}~\citep{vaswani2017attention} is an advanced model which achieves great performance in many NLP tasks.  

\item  \textbf{GatedNN} is an improved version of deep neural networks, where edges between neurons are gated by the gumble-softmax function so the network could be sparse. 
\end{itemize}

\noindent (2) The second type of baselines is to first learn a gene co-expression matrix as knowledge and then do the prediction. The co-expression matrix identifies which genes have a tendency to show a coordinated expression pattern, so it can be built by the similarity between gene expressions. We follow the process described in MLA-GNN \citep{xing2022multi} to learn the knowledge and then apply two classical graph neural networks, \textbf{GCN} \citep{kipf2017semi} and \textbf{GAT} \citep{velickovic2018graph}, to aggregate the information and do the prediction. 

\noindent (3) We also compare the models using the existing biological knowledge. This knowledge could be graphs, such as GRN or PPI.
\begin{itemize}
\item  \textbf{MLP+N2V} used node2vec \citep{grover2016node2vec} to learn node embeddings of the prior graph, combine them with gene expression data and employ MLP to predict. 
\item  \textbf{GCN} and \textbf{GAT} directly utilized the prior graph. As there are different prior graphs (GRN or PPI), we report the best results for GCN and GAT with the most suitable graph. 

\item  \textbf{DCell} \citep{ma2018using} used the Gene Ontology (GO) knowledge to build the neural network architecture. 
\item  \textbf{P-NET} \citep{elmarakeby2021biologically} utilized protein-pathway relations as prior knowledge and apply DNNs to predict. 
\item  \textbf{BFReg-NN} is our proposed method. There are basic and enhanced versions. 
\end{itemize}

\subsubsection{Training Details}
For optimizers, we use Adam for all the models. The hyperparameters are summarized in Table \ref{h_set} in Appendix.  
We set a small dimension of hidden embeddings to avoid overfitting. For the prediction module, we set the number of hidden units with the best value obtained from the validation data. 
In addition, the update function could be implemented by the sum or concatenation operation. We set $K=1$ for all the experiments.

The performance of BFReg-NN is influenced by the value of hyperparameter $\alpha$. While $\alpha$ controls the acceptance of newly discovered knowledge, it is dependent on whether the existing biological knowledge is suitable for the current dataset, as shown in Table~\ref{a_set} in Appendix. For example, in the future value forecasting task, $\alpha$ is larger than in other tasks to get the best result because existing knowledge is incomplete in the dataset. Besides, in the missing value task, our model performs better when $\alpha=0$ in the protein level, which means BFReg-NN ignores the irrelevant level when discovering knowledge. We vary it from \{0, 0.00001, 0.00005, 0.0001, 0.0005, 0.001, 0.005\} and select the best value when the loss is the smallest in the validation dataset.

\subsection{Main Results}
\label{sec:static_dynamic}
Here we describe the results from two aspects, static and dynamic. Missing gene expression prediction and cell classification are static tasks, while future gene expression prediction is dynamic tasks.
\begin{table}[t]
\setlength{\tabcolsep}{4pt}
\centering
\caption{Results on future gene expression value forecasting (evaluated by MSE$\downarrow$ and PCC$\uparrow$).}
\label{d_results}
\resizebox{0.5\textwidth}{!}{
\begin{tabular}{clll}
\toprule
\textbf{Knowledge}     & \textbf{Models}                              & \textbf{MSE}             & \textbf{PCC}             \\
\midrule
\multirow{5}{*}{\shortstack[c]{no prior\\knowledge}}               
     & MLP                                 & 0.0761 $\pm$    0.0038     & 0.9704  $\pm$ 0.0018       \\
     & LSTM                                & 0.0822   $\pm$ 0.0028      & 0.8888   $\pm$ 0.0034       \\
     & Transformer                         & 0.0743   $\pm$  0.0013      & 0.9717   $\pm$  0.0007      \\
     & Gated NN                            & 0.0743     $\pm$ 0.0026      & 0.9717    $\pm$ 0.0007      \\
     & Random Forest	                      &0.1115 $\pm$ 0.0021	        & 0.9685 $\pm$ 0.0006  \\
     & XGBoost	                          &0.0837 $\pm$ 0.0033	      & \textbf{0.9753} $\pm$ 0.0008   \\
\hline
\multirow{2}{*}{\shortstack[c]{co-expression}} & GCN                                 & 0.3838  $\pm$    0.0090     & 0.8293  $\pm$     0.0036    \\
              & GAT                                 & 0.3447   $\pm$    0.0236    & 0.8455      $\pm$  0.0114   \\\hline
\multirow{8}{*}{\shortstack[c]{existing\\knowledge}}              & MLP+N2V                             & 0.0760  $\pm$    0.0043   & 0.9706    $\pm$    0.0018   \\
              & GCN                                 & 0.0920  $\pm$   0.0053    & 0.9646   $\pm$    0.0020   \\
              & GAT                                 & 0.0869 $\pm$  0.0044        & 0.9661 $\pm$    0.0017       \\
              & BFReg-NN (MLP+B) & 0.0732      $\pm$  0.0026    & 0.9719  $\pm$   0.0009      \\
              & BFReg-NN (MLP+E)             & \textbf{0.0724} $\pm$ 0.0023 & 0.9724 $\pm$ 0.0007 \\
              & BFReg-NN (LSTM+B)            & 0.0825   $\pm$  0.0029     & 0.8906   $\pm$    0.0041   \\
              & BFReg-NN (LSTM+E)            & 0.0819  $\pm$    0.0027     & 0.8907    $\pm$   0.0049   \\
\bottomrule
\end{tabular}
}
\end{table}

\stitle{Static task results:} We report the main results for missing value prediction and cell classification, summarized in Table \ref{r_result} and Table \ref{c_results} respectively. Firstly, in general, compared to no prior knowledge methods, biological knowledge improves the performance of models effectively, which means that modeling specific meanings for neural networks could improve the performance in biological tasks. Although Gated NN includes the gumble-softmax function which also is able to predict edges in a probability, it is hard to generate the existing biological knowledge. Besides, co-expression knowledge does not show outstanding performance because it is extracted from a small size of data and thus brings the noise. Some complex model also achieves promising results, such as Transformer, which indicates that building dense biological relations may be needed for the cell samples. 

Secondly, BFReg-NN also outperforms the baselines using the existing knowledge. MLP+N2V fails in prediction because it uses the knowledge by embedding nodes implicitly, rather than in an explicit architecture. Compared with GCN and GAT, the basic BFReg-NN(B) could merge hierarchical knowledge from different levels of the biological system. Dcell and P-NET built the sketchy networks and applied DNNs to predict, which cannot simulate the complex gene/protein reactions. Thus, their performance are also worse than BFReg-NN(B).

Thirdly, the enhanced BFReg-NN(E) learns important undiscovered knowledge from the gene expression data, and enhances the performance further. Overall, BFReg-NN(E) achieves the best results in most cases.

\stitle{Dynamic task results:} Results on future value prediction are presented in Table~\ref{d_results}. 
Compared with directly employing MLP/LSTM to forecast the values, BFReg-NN followed with MLP/LSTM could improve the quality of embeddings and achieve better performance. GCN and GAT with co-expression matrix perform badly due to the small data size. When using existing knowledge, MLP+N2V, GCN and GAT still cannot obtain a good result because the knowledge is incomplete. For example, there are several isolated nodes in the dataset, and these methods fail in updating the embeddings of these nodes. Transformer and Gated NN merge the gene expression densely, so they obtain better results.

The best-performing models in this dataset are  tree-based ensemble methods, so we apply Random Forest and XGBoost. Random Forest has a poor performance on both MSE and PCC. XGBoost has a little higher value on PCC but also a higher MSE, which means XGBoost could predict the trend but fail on the specific values.

Finally, the enhanced BFReg-NN(MLP+E) could add the extra discovered knowledge to mitigate the problem of incomplete prior knowledge, and learn the embeddings with a suitable architecture. Therefore, it has the lowest MSE value and a higher PCC value closing to the best one. 

\begin{table}[t]
\centering
\caption{Performances of cell trajectory simulation (evaluated by WD$\downarrow$).} 
\label{cell_trajectory}
\resizebox{0.5\textwidth}{!}{
\begin{tabular}{lllllll}
\toprule
Model                & EGF             & iEGFR    & iMEK & iPKC & iPI3K & Avg.   \\
\midrule
MLP(S)            & 1.2829      & 1.2087 & 1.1471& 1.1320& 1.1362& 1.2265\\

MLP            & 1.0325      & 1.0592 & 1.0893& \textbf{1.0824}& \textbf{1.0827}& 1.0692\\

BFReg-NN (E)         & \textbf{0.8490} & \textbf{0.8753}& \textbf{0.9198}& 1.0830& 1.1077& \textbf{0.9670}\\
\bottomrule 
\end{tabular}
}
\end{table}

\begin{table}[t]
\centering
\caption{Performances of pre-trained BFReg-NN (evaluated by MSE$\downarrow$ and PCC$\uparrow$).} 
\label{pre_train}
\resizebox{0.5\textwidth}{!}{
\begin{tabular}{lll}
\toprule
Model                & MSE             & PCC    \\
\midrule
BFReg-NN (MLP+E)             & 0.0724     $\pm$  0.0023   & 0.9724$\pm$ 0.0007\\
Pre-trained BFReg-NN (MLP+E)  & \textbf{0.0719}$\pm$ 0.0042 & 0.9723$\pm$ 0.0018\\
\bottomrule 
\end{tabular}
}
\end{table}

\subsection{Results for Cell Trajectory}
\label{sec:cell_result}
To show the efficiency of the BFReg-NN model, we compare it with two baselines. MLP(S) indicates that genes do not interact with each other and the value is predicted by DNNs. MLP simply applied DNNs to aggregate all the genes and then simulate the trajectory one by one.
Since the number of situations is large, we show the result for five treatments averaged by different cell lines in Table~\ref{cell_trajectory}. In detail, the treatments include stimulating cells with EGF (epidermal growth factor), and four kinase inhibitors (iEFGR, iMEK, iPKC, iPI3K) combined with EGF. 

BFReg-NN obtained the small wessentrian distance compared with baselines on average. Specifically, BFReg-NN achieved significant results in EGF and iEGFR treatments. The reason is that gene value is constrained by their related genes during several time steps.
BFReg-NN considered that suitable relationship among genes, which helps to predict the trajectory for each gene. 
MLP(S) regarded each gene as independent, so it cannot use the other gene information to make a better prediction. MLP supposed that all the genes have a strong influence on each other, which is against the truth that some genes have more significant influence than others. Thus, MLP has a suboptimal performance.

\begin{figure*}[h!]
\begin{minipage}[h]{0.27\textwidth}
\centering
    \includegraphics[width=1\textwidth]{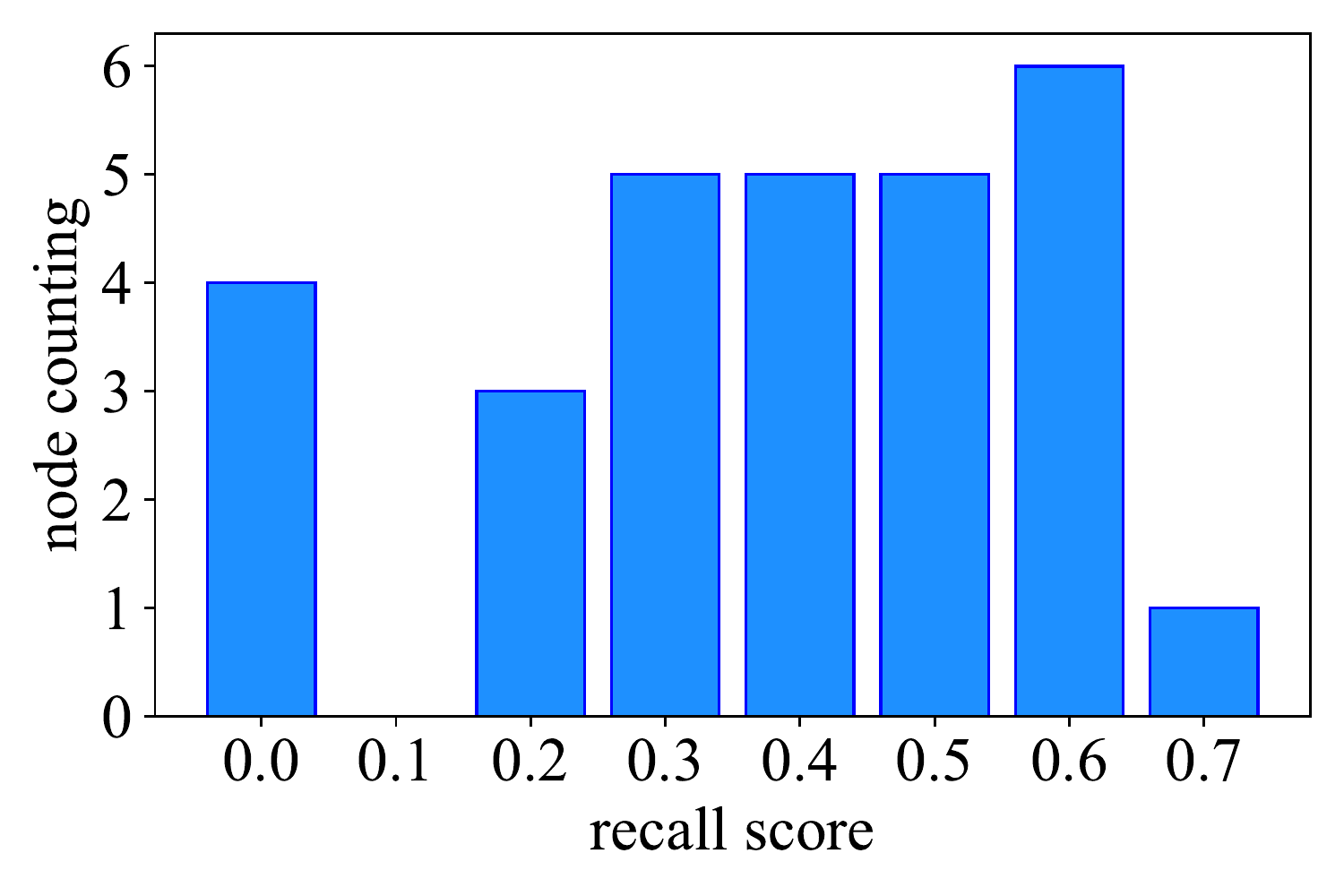}
\caption{Recall score for all the nodes in breast cancer. The average score is 0.4175.}
    \label{recall_fig}
\end{minipage}
\hspace{2mm}
\begin{minipage}[h]{0.4\textwidth}
    \centering
    \includegraphics[width=0.9\textwidth]{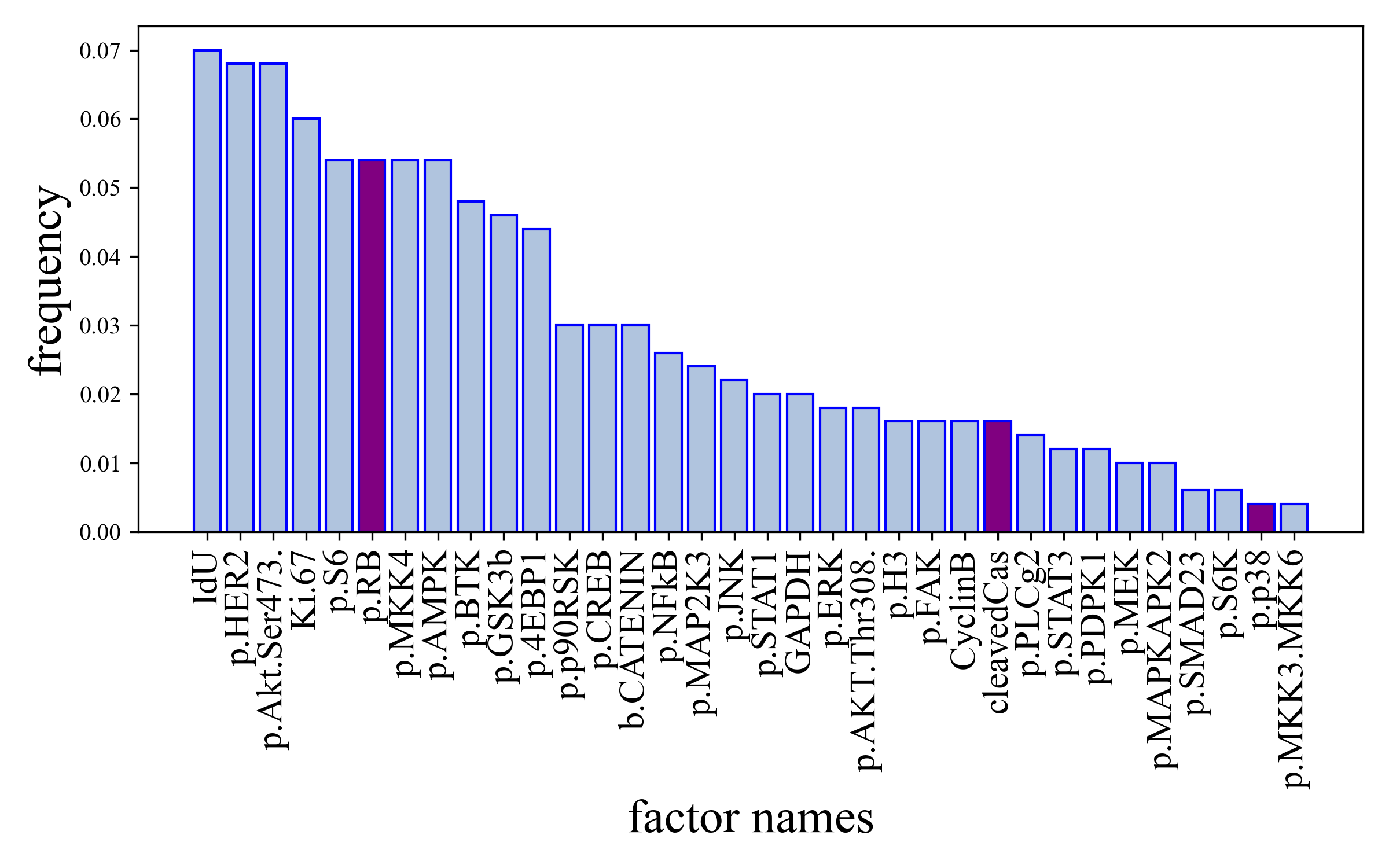}
    \vspace{-0.6em}
    \caption{Top-20 frequent interactions for gene p.p53}
    \label{figp53}
\end{minipage}
\hspace{2mm}
\begin{minipage}[h]{0.28\textwidth}
\centering
    \includegraphics[width=0.95\textwidth]{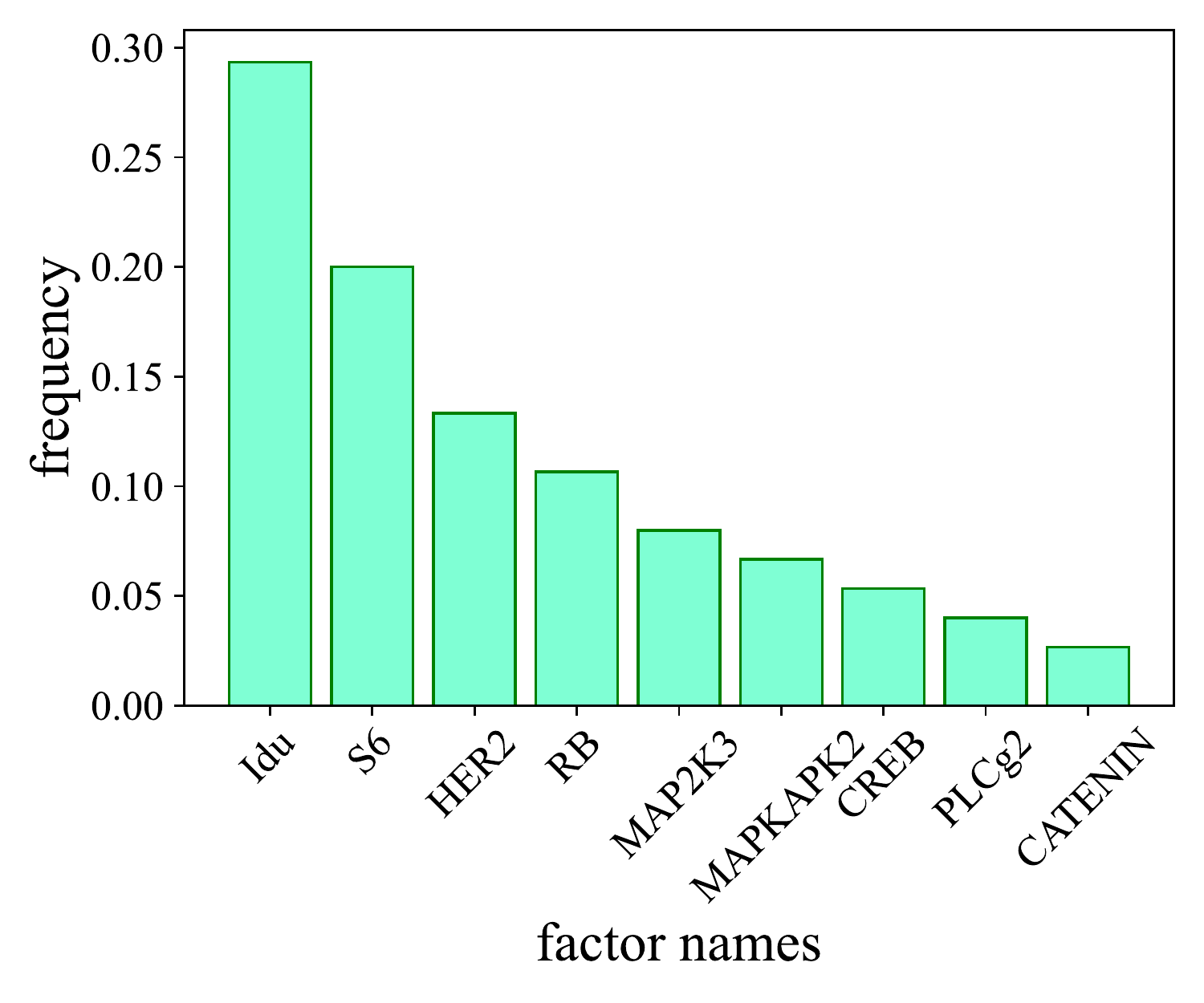}
    \vspace{-0.5em}
    \caption{Frequency of newly discovered interactions including p53.}
    \label{fre_fig}
\end{minipage}

\end{figure*}

\begin{table*}[h!]
\setlength{\tabcolsep}{2.8pt}
\caption{Ablation study results on missing value prediction (evaluated by MSE $\downarrow$).} 
\label{m_aresult}
\resizebox{0.8\textwidth}{!}{
\begin{tabular}{lllllllllll}
\toprule
Knowledge    & BT20           & HS578T         & LNCAP          & A549           & MCF7           & MCF10A         & MDAMB231       & PC3            & SKBR3          & A375           \\
\midrule
GRN          & \textbf{0.318}          & \textbf{0.317} & \textbf{0.320} & \textbf{0.297} & \textbf{0.308} & \textbf{0.306} & \textbf{0.320} & 0.335 & \textbf{0.307} & \textbf{0.308} \\
GRN\&PPI     & 0.343          & 0.342          & 0.338          & 0.326          & 0.339          & 0.333          & 0.343          & 0.357          & 0.338          & 0.344          \\
GRN\&Pathway & 0.335 & 0.335 & 0.326 & 0.314 & 0.324 & 0.319 & 0.331 & \textbf{0.332} & 0.327 & 0.325 \\  
 \bottomrule   
\end{tabular}
}
\end{table*}

\begin{table}[t!]
\centering
\caption{Ablation study results on cell classification (evaluated by macro AUC $\uparrow$).} 
\label{c_aresult}
\resizebox{0.5\textwidth}{!}{
\begin{tabular}{clllll}
\toprule
\multicolumn{1}{l}{\textbf{Models}} & \textbf{Knowledge} & \textbf{muscle} & \textbf{diaphragm} & \textbf{lung}   & \textbf{trachea} \\
\midrule
\multirow{3}{*}{B}    & GRN               & 0.8165          & 0.7564          & 0.8230          & 0.9376          \\
                          & GRN\&PPI          & 0.8798          & 0.8219          & 0.8784          & 0.9241          \\
                          & GRN\&PPI\&Pathway & 0.8808          & 0.8420          & 0.8808          & 0.9321          \\
\hline
\multirow{5}{*}{E} & GRN               & 0.8165          & 0.7873          & 0.8373          & \textbf{0.9446} \\
                          & GRN\&PPI          & 0.8807          & 0.8417          & 0.8892          & 0.9297          \\
                          & GRN\&PPI\&Pathway & \textbf{0.8884} & \textbf{0.8509} & \textbf{0.8903} & 0.9360          \\
                          & No Intra-level   & 0.8799          & 0.8382          & 0.8809          & 0.9235          \\
                          & No Inter-level   & 0.8790          & 0.8058          & 0.8268          & 0.9391         \\
 \bottomrule                                   
\end{tabular}
}
\end{table}

\subsection{Results for Pre-training}
\label{sec:pre-training_result}

BFReg-NN could benefit from the pre-training and fine-tuning framework, and the results are shown in Table ~\ref{pre_train}. The detailed experimental steps are: (1) Utilize the missing gene value prediction task to pre-train a BFReg-NN model on the breast cancer dataset; (2) Freeze the parameters of the BFReg-NN model except for the last MLP layers, and (3) Fine-tune the last MLP layers to do the future gene expression prediction. With the pre-training, the performance increases significantly on the MSE metric. Because biological knowledge is universal and similar in the cell across different tasks, it is reasonable that pre-trained BFReg-NN improves the other tasks’ performance with fine-tuning. However, the standard deviation slightly rises, which means a more stable method to fine-tune the model should be discussed in the future.

\subsection{Knowledge completion by BFReg-NN}
\label{sec:knowledge_result}

To demonstrate the ability of BFReg-NN to complete the knowledge, we evaluate it on the future value forecasting task. We conducted two type of experiments. One is to remove the existing edges and then see if these edges can be recalled by BFReg-NN. The other is to judge whether the newly discovered biological knowledge is also proved in the recent literature.

In the first experiment, for each gene, we remove its all the edges to make it an isolated node, and conduct the model to recover the knowledge. The discovered edges come out with weights, and we rank them to obtained the top-k list. We run BFReg-NN several times to avoid randomness. The frequency of an edge is computed by (the times that the edge is in the top-k list)/(the total runs). We select the top-$20$ interaction pairs with the highest frequency as discovered knowledge. The results are shown in Figure \ref{recall_fig}, where BFReg-NN achieves an average 0.4172 recall score for all the nodes. 

We also take p.p53 gene as an example. The detailed recalled edges are shown in Figure~\ref{figp53} where the purple color marks the edges existed in the current network. All the existing edges for p.p53 are recalled in the top-20 list. 

Secondly, some new knowledge is discovered by biological methods in recent breast cancer research, which is also found in our proposed BFReg-NN. We take the gene p53, a well-known tumor suppressor, as an example to verify the biological meaning of discovered knowledge. We show the top-$10$ frequent interaction pairs including p53 in Figure \ref{fre_fig}. And then we find the cues in recent biological literature. Iododeoxyuridine (Idu) is the most frequent marker, which identifies cell phases, indicating whether the cell division is continued in breast cancer~\citep{behbehani2012single, gabor2021cell}.  S6 takes the second place, which is regulated in mTOR signaling by p53, and the level of S6 increases when p53 is insufficient \citep{luo2021p53}. Besides, p53 also regulates the expression level of Her2, where the interaction frequency is about 0.15 in our prediction. As a kind of Her2-positive cancer, breast cancer could be treated by inhibiting p53 \citep{fedorova2020attenuation}. After that, the regulatory relation between p53 and RB is in the existing biological knowledge dataset.

Overall, we notice that most of the existing edges can be recalled by our model, while BFReg-NN also adds knowledge that does not appear in the current database.

\subsection{Ablation Study}
\label{sec:ablation_result}
In this section, we provide a brief description of the effectiveness of each part in BFReg-NN. The ablation analysis for the cell classification task is shown in Table \ref{c_aresult}. The accuracy is gradually improved when adding a higher level of knowledge to the model in most cases. In trachea, GRN shows the best performance which indicates PPI may lead to an overfitting problem in this organ. In addition, we present the results of \textit{no intra-level} and \textit{no inter-level} situations. 
\textit{No intra-level} version removes PPI and GRN knowledge and maintains the hierarchical network; and \textit{no inter-level} merges the PPI and GRN into a large graph and deletes the hierarchical structure. The results show that hierarchical structure is more important than simply linking the factors together, which obeys the phenomenon that the biological factors in a cell are produced step by step.

In addition, we could choose different levels from the biological system as layers for BFReg-NN to adapt to different tasks. For example, the cell classification task aims to capture the cell state involving a variety of factors in a cell, and thus it needs multiple levels of biological knowledge to build the cell phenotype. In contrast, shown in Table \ref{m_aresult}, the missing value task focuses on gene expression value, which means single gene level knowledge is enough while protein or pathway level knowledge may bring the noise to predictions.

\section{Conclusion}
In this paper, we propose a generic framework BFReg-NN, which offers biological meanings to neurons and links between neurons, and imitates the whole cell system as a neural network at both intra-level and inter-level. We apply BFReg-NN to different downstream tasks and our experimental results demonstrate that BFReg-NN consistently outperforms baselines and discovers new biological meaningful insights.
BFReg-NN provides a novel paradigm to merge cell sequencing data and biological knowledge, and it could be extended to more types of genomics data to simulate and understand complex biological systems in the future.

In the future, we plan to apply BFReg-NN in some real and important biological problems, such as predicting drug surgery effect for a special disease, and simulating stem cell development.

\bibliographystyle{ACM-Reference-Format}
\bibliography{ref}

\newpage
\appendix
\onecolumn

\section{Appendix}

\begin{table*}[ht]
\centering
\caption{Hyperparameter setting} 
\label{h_set}
\resizebox{0.7\textwidth}{!}{
\begin{tabular}{lllll}
\toprule
\multicolumn{1}{c}{\multirow{2}{*}{\textbf{Task}}} & \multicolumn{1}{c}{\multirow{2}{*}{\textbf{Learning rate}}} & \multicolumn{1}{c}{\multirow{2}{*}{\textbf{Max epoch}}} & \multicolumn{2}{c}{\textbf{Dimension of hidden embeddings}}       \\
\multicolumn{1}{c}{}                               & \multicolumn{1}{c}{}                                        & \multicolumn{1}{c}{}                                    & \textbf{BFReg-NN module} & \textbf{Prediction module} \\
\midrule
missing value prediction                                    & 1e-3                                                        & 200                                                     & 4                  & 1024                      \\
cell classification                                & 5e-4                                                        & 200                                                     & 4                  & 256                       \\
future value forecasting                           & 1e-4                                                        & 2000                                                    & 16                 & 512                      \\
 \bottomrule
\end{tabular}
}
\end{table*}

\begin{table*}[htb]
\centering
\caption{The setting of $\alpha$ in different tasks} 
\label{a_set}
\resizebox{1\textwidth}{!}{
\begin{tabular}{llllllllllll}
\toprule
\multicolumn{12}{c}{\textbf{Missing value prediction}}                                                                                                                                                                                                          \\
\midrule
\textbf{Knowledge}        & \textbf{$\alpha$ set} & \textbf{BT20} & \textbf{HS578T} & \textbf{LNCAP}     & \textbf{A375} & \textbf{A549}    & \textbf{MCF7}      & \textbf{MCF10A}  & \textbf{MDAMB231} & \textbf{PC3} & \textbf{SKBR3} \\
\midrule
GRN                       & $\alpha_1$            & 1e-2          & 1e-4            & 1e-5               & 1e-4          & 1e-3             & 0                                                     & 1e-3                          & 5e-4              & 1e-3         & 5e-5           \\
\multirow{2}{*}{GRN\&PPI} & $\alpha_1$            & 1e-4          & 0               & 5e-4               & 0             & 1e-4             & 5e-5                                                  & 5e-5                          & 5e-5              & 5e-4         & 5e-4           \\
                          & $\alpha_2$            & 0             & 0               & 0                  & 0             & 0                & 0                                                     & 0                             & 0                 & 0            & 0              \\
                           \bottomrule 
                          \toprule
\multicolumn{7}{c}{\textbf{Cell classification}}                                                                                        & \multicolumn{2}{c}{\textbf{Future value forecasting}} & \multicolumn{1}{c}{\textbf{}} &                               &                \\
\midrule
\textbf{Knowledge}        & \textbf{$\alpha$ set} & \textbf{GSE}  & \textbf{muscle} & \textbf{diaphragm} & \textbf{lung} & \textbf{trachea} & \textbf{Backbone}                                     & \textbf{$\alpha$ set}             &                   &              &                \\
\midrule
GRN                       & $\alpha_1$            & 1e-5          & 1e-5            & 5e-5               & 5e-5          & 5e-4             & MLP                                                   & 1e-2                          &                   &              &                \\
GRN\&PPI                  & $\alpha_1$            & 5e-4          & 1e-5            & 1e-5               & 1e-4          & 1e-4             & LSTM                                                  & 5e-3                          &                   &              &                \\
                          & $\alpha_2$            & 1e-5          & 0               & 5e-5               & 0             & 1e-5             &                                                       &                               &                   &              &               \\
 \bottomrule                          
\end{tabular}}
\end{table*}

\end{document}